%% file: paper_template.tex
\documentclass[conference]{IEEEtran}
\usepackage{times}

\usepackage[numbers]{natbib}
\usepackage{multicol}
\usepackage[bookmarks=true]{hyperref}
\usepackage{natbib}
\makeatletter
\newcommand*{\rom}[1]{\expandafter\@slowromancap\romannumeral #1@}
\makeatother
\usepackage{amsmath}
\usepackage[usenames,dvipsnames]{color} 

\usepackage{algpseudocode}

\algnewcommand\algorithmicnot{\textbf{not}}
\algdef{SE}[IF]{IfNot}{EndIf}[1]{\algorithmicif\ \algorithmicnot\ #1\ \algorithmicthen}{\algorithmicend\ \algorithmicif}%
\usepackage{algorithm}

\usepackage{graphicx}
\graphicspath{ {./images/} }
\usepackage{caption}
\usepackage{subcaption}

\usepackage{multirow}
\usepackage{booktabs} 

\begin{document}

\title{ Topology-Guided ORCA: Smooth Multi-Agent Motion Planning in Constrained Environments }





\author{\authorblockN{Fatemeh Cheraghi Pouria\authorrefmark{1}\authorrefmark{4},
Zhe Huang\authorrefmark{2}\authorrefmark{4},
Ananya Yammanuru\authorrefmark{3} \authorrefmark{4}, 
Shuijing Liu\authorrefmark{2} and
Katherine Driggs-Campbell\authorrefmark{2}}
\authorblockA{\authorrefmark{1} Department of Mechanical Science and Engineering\\
University of Illinois at Urbana-Champaign,
Urbana, Illinois 61801\\
Email: fatemeh5@illinois.edu}
\authorblockA{\authorrefmark{2} Department of Electrical and Computer Engineering\\
University of Illinois at Urbana-Champaign,
Urbana, Illinois 61801\\
Email: \{zheh4, sliu105, krdc\}@illinois.edu}
\authorblockA{\authorrefmark{3} Department of Computer Science\\
University of Illinois at Urbana-Champaign,
Urbana, Illinois 61801\\
Email: ananyay2@illinois.edu}
\authorblockA{\authorrefmark{4}denotes equal contribution as the first author}}


\maketitle
\input{Sections/0_abstract}

\IEEEpeerreviewmaketitle

\input{Sections/01-intro}
\input{Sections/01.5_related}
\input{Sections/02-method}
\input{Sections/03-exp}
\input{Sections/04-conclusion}

\section{Acknowledgment}
This work was supported by the National Science Foundation under Grant No. 2143435.


\bibliographystyle{plainnat}
\bibliography{references}

\end{document}

%% file: Sections/0_abstract.tex
\begin{abstract}
We present Topology-Guided ORCA as an alternative simulator to replace ORCA for planning smooth multi-agent motions in environments with static obstacles. Despite the impressive performance in simulating multi-agent crowd motion in free space, ORCA encounters a significant challenge in navigating the agents with the presence of static obstacles. ORCA ignores static obstacles until an agent gets too close to an obstacle, and the agent will get stuck if the obstacle intercepts an agent's path toward the goal. To address this challenge, Topology-Guided ORCA constructs a graph to represent the topology of the traversable region of the environment. We use a path planner to plan a path of waypoints that connects each agent's start and goal positions. The waypoints are used as a sequence of goals to guide ORCA. The experiments of crowd simulation in constrained environments show that our method outperforms ORCA in terms of generating smooth and natural motions of multiple agents in constrained environments, which indicates great potential of Topology-Guided ORCA for serving as an effective simulator for training constrained social navigation policies.







\end{abstract}

%% file: Sections/01-intro.tex
\section{Introduction}
There have been groundbreaking advancements in social navigation in recent years thanks to the reinforcement learning scheme with effective simulators for planning crowd motion~\cite{liu2023intention}. In particular, Optimal Reciprocal Collision Avoidance (ORCA) has been prevalent for multi-agent motion planning in crowd simulation due to their robust capabilities to compute collision-free trajectories for a large number of agents~\cite{van2011reciprocal, chen2017socially, chen2019crowd}. However, many social navigation algorithms are investigated in free open space setting, which does not hold in many real-world applications, since many mobile robots serve in indoor environments, for example restaurants and hotels. We argue that this is due to the fact that ORCA, as one of the predominant human agent policies in simulation, is not able to effectively handle constrained environments with static obstacles. The agents taking ORCA as their motion policy suffer from a lack of feasible and smooth trajectories to move around static obstacles in constrained environments \cite{perez2021robot}. If any static obstacle intercepts an agent's path, there is a chance that the agent does not sense the presence of the obstacle until it gets too close to be able to move around it.

To address this limitation, we introduce Topology-Guided ORCA, which uses path planning in constrained environments to guide ORCA policy. We apply Medial Axis Transform to generate a topological graph of the traversable region in the constrained environment. With respect to each agent, we augment the graph by finding the collision-free edges between the nodes of the topological graph and the agent's start and goal positions. We use the augmented graphs to perform path planning to generate waypoints which serve as a sequence of goals for ORCA policy of the agent. We conduct crowd simulation experiments in constrained environments of different settings. We show that Topology-Guided ORCA consistently outperforms ORCA, and exhibits natural and reasonable crowd motion with the presence of static obstacles. We believe that Topology-Guided ORCA is a promising multi-agent motion planning method which can provide high-quality simulation data for training social navigation policies in constrained environments.

%% file: Sections/01.5_related.tex
\section{Preliminaries}



Velocity Obstacle (VO) methods have attracted broad attention over decades due to their ability to effectively plan multi-agent motions. VO was first introduced by \citet{fiorini1998motion} to define a first-order approximate set of robot's velocities that may result in a collision during the navigation time horizon. As a result, moving agents should select velocities outside of that set to ensure collision-free navigation.

As an extension of VO, \citet{van2008reciprocal} developed the Reciprocal Velocity Obstacles (RVO) method to tackle the velocity oscillation problem when two agents move toward each other with opposite velocities. In addition, RVO only considers agents and obstacles within a predefined neighbor region to boost computation speed. 


Developed along this line of work, Optimal Reciprocal Collision Avoidance (ORCA) addresses multiple decision-making entities\citet{van2011reciprocal}. ORCA solves a low-dimensional linear program problem to find agents' collision-free velocity. They define an agent's preferred velocity as the velocity at which the agent reaches its destination, given that no other agents are in the way. The issue with this preferred velocity is that it is an internal state and not observable by other agents.
Thus, an optimization problem with convex constraints should be solved to find the closest possible velocity to the preferred one. Constraints are generated from the intersection of 2-D planes demonstrating admissible velocities for each pair of agents.

According to  \citet{van2011reciprocal}, for a given agent $A$, ORCA solves the following optimization problem to establish $A$'s velocity as close as possible to its preferred velocity:
$$v^{new}_A = \operatorname*{argmin}_{v \in ORCA^\tau_A } \|v-v^{pref}_A\|$$

When two moving agents which are following the same motion policy get close to each other, the optimization problem presented above is guaranteed to have feasible solutions for both agents. 
Hence, the agents deviate from each other and no collision occurs. 
However, when dealing with static obstacles, the moving agent is the only one responsible for adjusting the velocity and avoiding collision. \citet{van2011reciprocal} states that the complement of the velocity obstacle for the agent and static obstacle is non-convex, which disallows us to apply linear programming. Therefore, the agent will not collide with the obstacle but there is a chance that it gets stuck behind it and does not move around.


 
This problem is also mentioned by \citet{perez2021robot}. They indicate that despite the advantages of using ORCA to simulate crowded and multi-agent scenarios, it suffers from not being able to provide trajectories that require a turn, such as going around a static obstacle. Given that, getting stuck behind the obstacle wall or encountering concave obstacles makes the goal invisible and results in agent's inability to navigate through the environment. \citet{perez2021robot} overcomes this limitation by decomposing the environment into specific layouts and regions of initial and goal positions, allowing the ORCA to navigate agents within these predefined layouts. However, this approach generates feasible trajectories by ORCA; the freedom of agents to choose their start and goal positions arbitrarily anywhere in the traversable environment is restricted.  
Another drawback of ORCA is revealed when a large number of agents encounter or get close to each other. Due to the dense environment and linear constraints that make ORCA more conservative, agents' velocities might drop or no feasible solution might be found. \citet{arul2021v} proposed V-RVO method that tends to accelerate the performance in such environments. To alleviate mentioned problems we aim to use a topological graph to represent the environment and guide ORCA as our motion planner.

%% file: Sections/02-method.tex
\section{Methodology} \label{sec:method}

ORCA agents can get stuck when a static obstacle is on their straight paths toward their goals, as ORCA may ignore the obstacle until the agents move too close to be able to turn around. Thus, we propose Topology-Guided ORCA to direct ORCA agents to move around static obstacles. In contrast to ORCA which uses fixed goals for the agents, Topology-Guided ORCA plans a path between start and goal positions of the agents, and sets the goal of the agents as the path waypoints in sequence. To plan an effective path for agent motion guidance, we represent the traversable region of the environment with a graph composed of two parts: a topological part which only depends on the environment, and an augmentation part which is agent-specific. 





\begin{figure}
     \centering
     \begin{subfigure}{0.2\textwidth}
         \centering
         \includegraphics[scale=0.2]{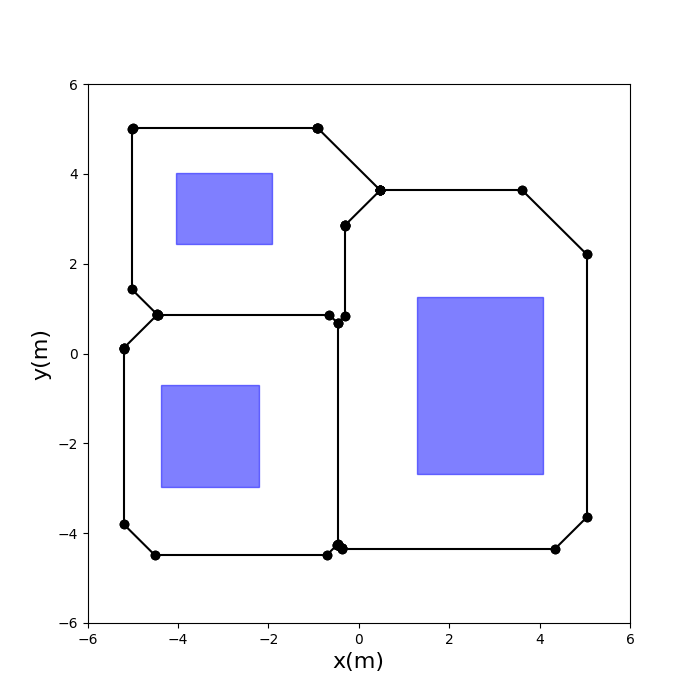}
         \caption{Topological graph}
         \label{fig:plain skeleton}
     \end{subfigure}
     \begin{subfigure}{0.2\textwidth}
         \centering
         \includegraphics[scale=0.2]{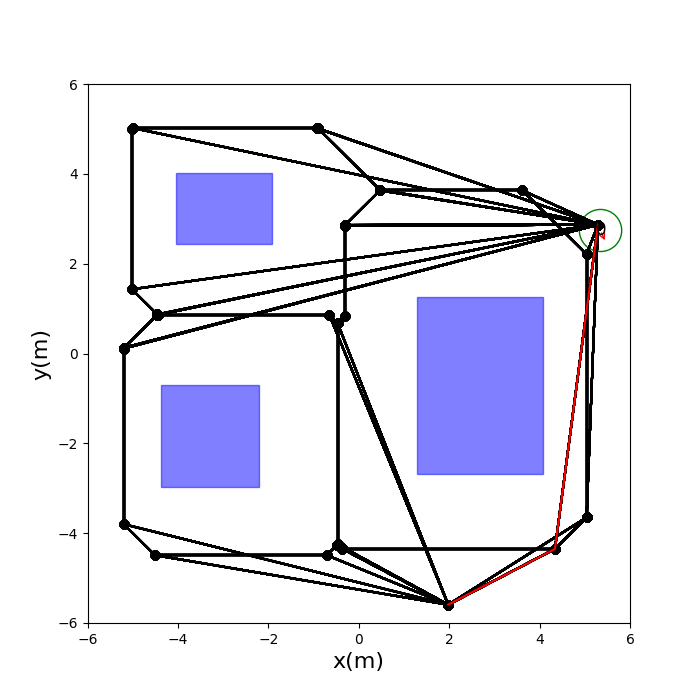}
         \caption{Augmented graph}
         \label{fig:augmented graph}
     \end{subfigure}
     \caption{(a) The graph representation of an environment including three static obstacles and (b) An agent's augmented graph. The green circle is the agent, and red lines show its path toward the goal. As one can see, the generated path guides the agent to move around the obstacles.}
        \label{fig:three graphs}
\end{figure}

\textbf{Topological Graph.} 
To help navigate agents around the obstacles, a topological graph needs to be built to reflect the connectivity of the free space, and to accurately represent all homotopic paths around the obstacles. Additionally, the speed of topological graph construction should meet the needs for online planning. In this work, we construct the topological graph $G = (V, E)$ with a Medial Axis graph \cite{Zhang1984AFP}, of which an example is presented in Figure \ref{fig:plain skeleton}. 




\textbf{Augmented Graph.}  At the beginning of the simulation, each agent is assigned an arbitrary start $s$ and goal $g$ position. 
Once an agent's start and goal positions are set, 
we augment the topological graph for each agent individually by adding $\{s,g\}$  
and draw edges $\{(u, v) | u \in \{s, g\}, v \textrm{ is visible from u} \}$ (see Figure \ref{fig:augmented graph}). 
In this context, a visible node refers to a node not occluded by obstacles that can be connected to the start or goal using a straight line. The motivation for having a separate graph for each agent is that ORCA usually selects a straight path toward the goal, assuming no other moving or static agents intercept the path. 
Each agent's graph is constructed such that if no obstacle is in the middle of the straight path from $ s$ to $ g$, an edge connecting $s$ to $g$ to the graph is added.


\textbf{Planning for Guidance.} We perform a shortest-path search from $s$ to $g$ on the augmented graph to generate waypoints. Waypoints are a sequence of augmented graph nodes, with $s$ as the first node and $g$ as the last node. 
We then use ORCA to navigate the agent between waypoints in sequence. Once an agent reaches its goal, a new goal is chosen, a new augmented graph is constructed, and the process is repeated. 


%% file: Sections/03-exp.tex
\section{Results and discussion} \label{sec:results}
\subsection{Simulation environment}
Our simulation setup includes 4 moving agents and 3 rectangular obstacles. Considering agents' radius and a margin equal to agents' radius around obstacles, 80 Percent of the area in the whole environment is traversable. 

This paper aims at addressing the limitations of ORCA in environments with static obstacles. To clearly illustrate the difference between our method and ORCA, we first focus on static obstacles in an environment with fewer agents. Static obstacles are randomly generated. Moreover, we investigate the same obstacle arrangement with 10 agents.      

The following subsection provides an overview of the metrics we consider to evaluate our method. We evaluate these metrics on 200 episodes, each consisting of 196 frames.

\subsection{Evaluation metrics}
To evaluate the performance of our proposed method, we designe a filter that takes into account agent velocities and outputs an array including each agent's status in each frame. An agent's status in a frame demonstrates whether it is getting stuck or moving naturally. A significant drop in velocity in the middle of the way is interpreted as getting stuck, which can be caused by either getting too close to static obstacles or moving in a crowded area. Moreover, the frame number in which an agent reaches its goal is also recorded. We quantify the filter output with the following metrics:
\begin{enumerate}
  \item Average agent velocity per path:  An agent's velocity is defined as the total distance traveled by the agent until it reaches its goal, divided by the number of frames required to complete this path. To calculate this metric, we average all agents' velocities across all episodes.
  \item Average percentage of mutual frozen frames per episode: A frame is called frozen if the agent is stuck. This metric computes the percentage of mutual frozen frames (i.e., all agents are stuck in the same frame) per episode and averages them across all episodes. 
  \item Average percentage of frozen frames per path: This metric computes the percentage of frozen frames per each agent's path and averages them across all episodes.
  \item Average number of occupied paths out of average number of paths per episode:  An occupied path is defined as a path including 30 consecutive frozen frames, equivalent to being frozen for more than 15 percent of an episode. This metric calculates the number of occupied paths and the total number of paths for each episode and averages them across all episodes. 
  \item Percentage of stuck agents: A stuck agent is defined as an agent that has not reached its goal from the beginning of the episode till the end and has a velocity less than one-third of average velocity. This metric computes the percentage of stuck agents out of the total number of agents across all episodes.
\end{enumerate}
These metrics are computed over each path from start to goal. Once an agent reaches its goal, its index is updated, and metrics are computed again for the new path toward the new goal.


\subsection{Results}
\begin{table*}[!h]
\begin{center}
\begin{tabular}{l *4c}
\toprule
Metrics &  \multicolumn{2}{c}{4 agents} & \multicolumn{2}{c}{10 agents}\\
\midrule
{}   & MA   & ORCA    & MA   & ORCA\\
\cmidrule{2-5}
Agent velocity $\uparrow$  &  17.66 & 15.48  & 15.34  & 13.9\\
\% of mutual frozen frames per episode $\downarrow$  &  7.92  & 15.03   & 17.29 & 23.65\\
\% of frozen frames per path $\downarrow$  & 6.05 &  8.54    &12.39 & 14.38\\
\# of occupied paths out of $\downarrow$ \# of paths  per episode $\uparrow$  &0.27/19.11 & 1.15/17.27   & 2.84/35.45  & 4.63/33.12\\
\% of stuck agents $\downarrow$   & 0.05 &  1.74    & 0.28 & 2.85\\
\bottomrule
\end{tabular}
\caption{quantitative metrics for 4 and 10 agents over 200 episodes}\label{table-quantitative}
\end{center}

\end{table*}

Quantitative results of our algorithm and ORCA are reported in Table \ref{table-quantitative}. The results show the advantages of using an environment's graph to guide ORCA in both scenarios with different numbers of agents. However, planning through a more crowded environment, such as having 10 agents instead of 4, worsens metrics for both methods, Our methods still perform better than ORCA in a more crowded scenario.

In general, as the average agent velocity shows, agents following our framework acquire higher velocities, indicating that their paths are less intercepted. According to the average percentage of mutual frozen frames per episode, our method reduces the likelihood of all agents being stuck in the same frame. Considering the average percentage of frozen frames per path, agent paths also include fewer frozen frames, meaning their velocities drop less often. Based on the average number of occupied paths out of average number of paths per episode and the percentage of stuck agents, our proposed method results in fewer occupied paths and fewer stuck agents per episode. A lower number of stuck agents indicates a higher probability of completing paths within a single episode as the total number of paths in the fourth metric shows. 

Overall, our method performs well in a four-agent environment, as the average velocity is higher and all other metrics are lower compared to ORCA. The only issue of ORCA in a not crowded environment is the presence of static obstacles. Thus, we show agents following our method are able to move around obstacles more efficiently compared to ORCA. Regardless of the motion planning method, more crowded environment leads to lower average velocity and higher metrics, which is reasonable since agents encounter more often and need to adjust their velocity to pass by each other without collision. Nevertheless, our proposed method surpasses ORCA in a crowded environment as well.

%% file: Sections/04-conclusion.tex
\section{Conclusion and future work} 
ORCA can simultaneously navigate multiple agents toward their goals in a real-time and collision-free manner. However, challenges arise regarding environments with static obstacles, which prevents ORCA from being effective in simulating crowd motion in constrained environments. The major problem occurs in the presence of static obstacles when ORCA is the only motion policy responsible for motion planning. In this case, the ORCA agent will likely get stuck behind an obstacle while moving toward its goal position. This paper alleviates this problem by guiding ORCA agents using the environment topology. The constrained crowd simulation experiments show that agents following our Topology-Guided ORCA maintain higher velocities and fewer frozen frames moving towards their goals, and occupied paths and stuck agents are less likely to happen, which indicates great potential for simulating constrained crowd motion and for training constrained social navigation policies.

In future work, this method can be integrated into more complicated environment setups, such as increasing the obstacle density or adding polygon or concave obstacles, hallways, and doors to the environment. We plan to investigate the effects of using probabilistic roadmap (PRM) as another topological graph construction method and exploring how this representation of traversable environment can help ORCA during navigation.
